\documentclass[english,11pt]{article}

\usepackage{amsmath, amssymb}
\usepackage{graphicx}
\usepackage{hyperref}
\usepackage[linesnumbered,ruled]{algorithm2e}
\usepackage{color}
\usepackage{subfig}
\usepackage{fullpage}

\hypersetup{colorlinks=true,linkcolor=blue,filecolor=blue,urlcolor=blue,citecolor=blue}

\begin{document}

\title{Sustainable Cooperative Coevolution with a Multi-Armed Bandit}

\author{Fran\c cois-Michel~De~Rainville$^\star$, Mich\`ele~Sebag$^\dagger$, Christian~Gagn\'e$^\star$,\\Marc~Schoenauer$^\dagger$, and Denis~Laurendeau$^\star$ \vspace{0.25cm}\\
{\small\begin{tabular}{cc}
	$^\star$Laboratoire de vision et syst\`emes num\'eriques			&
	$^\dagger$\'Equipe TAO, CNRS -- INRIA -- LRI 				\\
	D\'epartement de g\'enie \'electrique et de g\'enie informatique	&
	Universit\'e Paris-Sud						\\
	Universit\'e Laval, Qu\'ebec (Qu\'ebec), Canada~~G1V 0A6		&		
	F-91405 Orsay Cedex, France\\
\end{tabular}}
}

\date{}

\maketitle
\begin{abstract}
This paper proposes a self-adaptation mechanism to manage the resources allocated to the different species comprising a cooperative coevolutionary algorithm. The proposed approach relies on a dynamic extension to the well-known multi-armed bandit framework. At each iteration, the dynamic multi-armed bandit makes a decision on which species to evolve for a generation, using the history of progress made by the different species to guide the decisions. We show experimentally, on a benchmark and a real-world problem, that evolving the different populations at different paces allows not only to identify solutions more rapidly, but also improves the capacity of cooperative coevolution to solve more complex problems.
\end{abstract}

\noindent\textbf{\large Categories and Subject Descriptor}\\
I.2.8 [\textbf{Artificial Intelligence}]: Problem Solving, Control Methods, and Search --- \textit{heuristic methods}

\section{Introduction}
\label{sec:Introduction}

Coevolutionary algorithms~\cite{Axelrod1987,Hillis1990} are extending traditional evolutionary algorithms by making use of several subpopulations of individuals, often designated as species, and evaluating the fitness of the individuals jointly with individuals from other species. Cooperative coevolutionary algorithms~\cite{Potter2001} exploit modularity to decompose a problem into multiple interacting subcomponents, in a divide-and-conquer fashion. It is a paradigm where subcomponents not only emerge naturally from the evolution, but also cooperate to solve much more complex problems than those that can be tackled using standard single-species genetic algorithms. Examples of successful applications of cooperative coevolution include ensemble pattern classification~\cite{GarciaPedrajas2005}, multi-agent learning~\cite{Panait2005}, and sensor placement~\cite{DeRainville2012}, among others.

Fair competition between the subcomponent populations has been introduced in the original algorithms. Indeed, each population can run for an arbitrary number of generations before the other populations provide new cooperating individuals. Potter and De Jong~\cite{Potter2001} present an asynchronous evolution of the populations, although only few have reported making their optimization in an asynchronous manner. However, for situations where the natural decomposition of the problem yields subproblems of varying difficulty, it is natural to assign a different number of generations to each subpopulation. The issue is then to determine how the generations should be allocated to each species.

That issue can be formulated as a decision problem where we need to determine which population should be evolved at each time step, in order to maximize the global progress of the coevolutionary algorithm. The multi-armed bandit framework~\cite{Gittins1979,Lai1985}, widely studied in the context of game theory, strives at proposing actions maximizing the expected reward while acquiring more knowledge on the decision process. In our context, it is thus a natural and promising avenue to follow for an automatic subpopulation selection mechanism. Therefore, we propose to apply a dynamic multi-armed bandit to allocate more computational time to the most promising species in order to sustain the coevolution. For this purpose, we also present an efficient strategy to evaluate the rewards, in order to balance the evolution between the species, as this is a non-stationary decision process. 

The structure of the paper is as follows. First, in Sec.~\ref{sec:coopcoev} and \ref{sec:mab}, we discuss the important concepts involved in cooperative coevolution and multi-armed bandits, respectively. Then, we present our framework for the selection of the most promising population in Sec.~\ref{sec:popsel}. Next, we present experiments with our framework in Sec.~\ref{sec:Experiments} on an artificial problem and a real-world scenario on sensor placement. Finally, we conclude the paper on the contributions of this article and possible future work in Sec.~\ref{sec:Conclusion}.

\section{Cooperative Coevolution}
\label{sec:coopcoev}

Cooperative coevolution has been proposed by Potter and De Jong~\cite{Potter2001} to exploit the modularity of a problem, similarly to the rules in Learning Classifier Systems and subroutines in Genetic Programming. In this paradigm, each subpopulation, the so-called species, represents a subcomponent of a solution, and a complete solution is obtained by combining a representative of each species. Each individual is assigned a fitness equal to the reward of the complete solution in which it participates. The number of species can change along the evolutionary process, new species being introduced when the evolution stagnates and species being removed when their best individual does not contribute enough to the complete solution. Potter and De Jong also propose to evolve the species asynchronously, which would allow to allocate more time to some species. To the authors knowledge, no work has been made to adjust the effort on each species to sustain the cooperative optimization. However, multiple avenues have been explored to improve the original algorithm, some are presented in this section.

A multiple interactions scheme for the cooperative coevolution has been studied by Wiegand~\cite{Wiegand2003} to increase the convergence of the algorithm, but it has been shown to generate subcomponents that are over-generalizing to fit to a multitude of collaborators.
Furthermore, Bucci and Pollack~\cite{Bucci2005} proposed a modification of the original cooperative coevolutionary algorithm by replacing the selection mechanism with a new one based on Pareto dominance. However, establishing the Pareto dominance among the individuals of each species requires a tremendous number of fitness evaluations when the population size or the number of populations increases. In fact, it requires $O(nm^{k-1})$ evaluations, where $n$ is the number of individuals in the evaluated species, $m$ is the number of individuals in each population with whom each individual is evaluated (required for evaluating Pareto dominance), and $k$ is the number of populations. Such solution is not always possible for real world problems and thus other avenues need to be explored.

Panait et al.~\cite{Panait2006} biased the coevolutionary process by augmenting the fitness of an individual with an estimate of its best possible reward if it was evaluated with its optimal collaborators. The problem is now how to find the optimal collaborators for an individual in a non trivial problem. The authors propose to pair the individuals with the most successful collaborators found yet; or with collaborators that have cooperated well with structurally similar individuals; among other ideas. To our knowledge, none of these options have been tested nor compared with their proposed method when the optimal collaborators are known, making this solution unapproachable for real world problems.

Wu and Banzhaf~\cite{Wu2010} introduced a hierarchical decomposition of the problem domain. Cooperation among the individuals of the population is rendered by ``groups'' that are evolved in the framework. Groups provide a collaborating mechanism by assembling different individuals together into a solution which can be evaluated. In their algorithm, individuals and groups have their own fitness, which is shared by all individuals in a sharing radius to preserve diversity in the populations. Their technique exploits the uniqueness of the contribution of each subcomponent in a group to avoid carrying ``free riders'' that would have the same effect as bloat in Genetic Programming. Such uniqueness measures might not always be available for real world problems.

Albeit the promising enhancements to the original algorithm, this work is based on the standard cooperative coevolution of Potter and De Jong~\cite{Potter2001}, for its simplicity, its general applicability to any problems, and its ease of implementation. The pseudo-code is presented in Algo.~\ref{algo:coev}. 
\begin{algorithm}[t]
	\SetAlgoVlined
	\DontPrintSemicolon
	initialize $\mathfrak{P} \leftarrow \{\mathfrak{S}_i, i=1, \ldots, n\}$\;
	choose $\mathfrak{R} \leftarrow \{\operatorname{select\_random}(\mathfrak{S_i}), i=1, \ldots, n\}$\;
	\While{$\neg$stop}{ \label{line:while}
		\ForEach{species $\mathfrak{S}_i \in \mathfrak{P}$}{\label{line:for}
			$\mathfrak{S}_i \leftarrow \operatorname{apply\_variations}(\mathfrak{S}_i)$\;
			evaluate($\mathfrak{S}_i, \mathfrak{R}\backslash\mathbf{r}_i$) \label{line:evaluate}\; 
			$\mathfrak{S}_i \leftarrow \operatorname{select}(\mathfrak{S}_i)$\;
		\label{line:endfor}}
		$\mathfrak{R} \leftarrow \{\operatorname{select\_best}(\mathfrak{S_i}), i=1, \ldots, n\}$\;
		
		\If{improvement $< T_\mathrm{i}$}{\label{line:improvement}
			remove species with \textit{contribution} $< T_\mathrm{c}$\;
			$\mathfrak{R} \leftarrow \mathfrak{R}\backslash\mathfrak{R}^-$\;
			add a new species $\mathfrak{P} \leftarrow \mathfrak{P} \cup \{\mathfrak{S}^\prime\}$\;
			$\mathfrak{R} \leftarrow \mathfrak{R} \cup \{\operatorname{select\_random}(\mathfrak{S}^\prime)\}$\;
		}\label{line:endimprovement}
	}
	\caption{Cooperative co-evolution.}
	\label{algo:coev}
\end{algorithm}
First, a population $\mathfrak{P}$ of $n$ species $\mathfrak{S}_i$ is initialized randomly and one representative $\mathbf{r}_i$ of each species is placed  in the representative set $\mathfrak{R}$. Then, lines~\ref{line:for} to~\ref{line:endfor} show how each species is evolved independently of the others and how interaction through evaluations is achieved. More specifically, on line~\ref{line:evaluate}, the evaluation of the individuals of a species is made in collaboration with the  representatives of the other species, thus excluding the representative of the current species from set $\mathfrak{R}$ ($\mathfrak{R}\backslash\mathbf{r}_i$). The inner loop, beginning at line~\ref{line:for}, is a classical genetic algorithm, which can be replaced by any other evolutionary algorithm. At each ecosystem generation, i.e. the outer loop beginning at line~\ref{line:while}, the representatives are updated with the best individual of each species for the next generation. Then, the improvement of the evolution is verified at line~\ref{line:improvement}. If the solution fitness, given by all representatives, has not improved over the last $I$ generations by more than a given threshold $T_\mathrm{i}$, then the system is considered as stagnating. In this situation, unproductive species that contribute less than a threshold $T_\mathrm{c}$ are removed from the population
\footnote{Newer species are removed first and the contribution is re-calculated after each removal.}
-- their representatives ($\mathfrak{R}^-$) are also removed from $\mathfrak{R}$ -- and one new species is added to the population. The evolution continues until the termination criterion is reached.

\section{Multi-Armed Bandit}
\label{sec:mab}

Multi-Armed Bandits (MAB), introduced by Robbins~\cite{Robbins1952}, are used to model the exploration-exploitation trade-off faced by an agent that  takes actions in a given environment for which rewards are issued. In this setting, the agent has two opposite goals, which are: 1) to gain more knowledge about the environment and the rewards obtained (exploration), and 2) to exploit its current knowledge on the actions in order to maximize its immediate reward (exploitation). The standard MAB problem consists in a set of $k$ arms, each arm representing a possible action. A fixed and unknown reward probability distribution $r_i\in[0, 1]$ is associated to each arm. At each time step of the problem, the agent selects an arm and receives a reward according to the corresponding reward probability.

Auer et al.~\cite{Auer2002} proposed the Upper Confidence Bound (UCB) algorithm, which has been demonstrated to achieve asymptotically the optimal regret rate in the classical MAB case of independent and stationary reward probability distributions. This means that this algorithm strives at finding the best arm as fast as possible, while trying the other arms at an exponentially decreasing frequency. In an environment involving coevolutionary algorithms, none of the independent and stationary reward distribution assumptions are respected. Indeed, the quality of solutions in a species strongly depends on the solutions of the other species, which also makes the probability distributions non-stationary.

Da Costa et al.~\cite{DaCosta2008} proposed a dynamic MAB combining the principles of the UCB and the Page-Hinkley statistics for restarting the method. This bandit was used to automatically select variation operators during the course of an evolutionary algorithm, with promising results. Fialho et al.~\cite{Fialho2010} refined this dynamic MAB in order to improve the robustness and invariance to monotonic transformations. They added a rank-based credit assignment scheme based on the Area Under the Curve (AUC) measure, commonly used in machine learning.

Fialho et al.~\cite{Fialho2010} AUC dynamic MAB has been used as a basic species selection mechanism for the current work. We slightly modified the credit assignment by not taking into account the ties in the ranked rewards, since our reward model is binary ($\rho \in \lbrace0, 1\rbrace$) instead of continuous (see Sec.~\ref{sec:popsel}). The order of appearance of the rewards is thus taken into consideration, as a more recent reward obtains a higher credit than an equal but older one. Algo.~\ref{algo:auc} presents the credit assignment scheme, with $d\in [0,1]$ being the decay factor proposed in~\cite{Fialho2010} and $\mathbf{w} = [w_1~\cdots~w_n]$ contains the tuples rewards, index obtained by the arms in the last $n$ iterations.

\begin{algorithm}[t]
	\SetAlgoVlined
	\DontPrintSemicolon
	\KwIn{arm index $i$, window of rewards $\mathbf{w}$}
	\KwOut{credit $q$}
	sort lexicographically $\mathbf{w}$ according to rewards (best first), and ages in case of ties (newer first)\;
	$q \leftarrow y \leftarrow 0$\;
	\ForAll{ranks $r \in [1~\cdots~|\mathbf{w}|]$}{
		$\rho, j \leftarrow w_r$\;
		$\rho \leftarrow d^{(r-1)}(|\mathbf{w}| - (r -1))$\;
		\uIf{i = j}{
			$y \leftarrow y + \rho$\;
		}
		\Else{
			$q \leftarrow q + y\rho$\;
		}
	}
	\caption{Area under the curve credit assignment.}
	\label{algo:auc}
\end{algorithm}

The selection mechanism of the MAB is the same as the one proposed in \cite{Fialho2010}, it is presented in Algo.~\ref{algo:mab}.
\begin{algorithm}[t]
	\SetAlgoVlined
	\DontPrintSemicolon
	$\mathbf{n} \leftarrow [0~\cdots~0], \mathbf{q} \leftarrow [0~\cdots~0], \mathbf{w} \leftarrow []$\;
	\While{$\neg$stop}{
		\eIf{$\exists n_i \in \mathbf{n} : n_i = 0$}{\label{line:select}
			$\mathit{arm} \leftarrow \operatorname{select\_random}(\lbrace i | n_i = 0\rbrace)$\;
		}{
			$\mathit{arm} \leftarrow \arg \max_i \left(q_i + C\sqrt{\frac{2\log\sum_{j=1}^{n} n_j}{n_i}}\right)$\;\label{line:endselect}
		}
		apply selected $\mathit{arm}$ and retrieve $\mathit{reward}$\;
		$\mathbf{w} \leftarrow [(\mathit{reward}, \mathit{arm})~\mathbf{w}]$\;
		\If{$|\mathbf{w}| > W$}{
			$\mathbf{w} \leftarrow [w_i\in\mathbf{w}, i=0, \ldots, W-1]$\;
		}
		$\mathbf{n} \leftarrow [\operatorname{count}(i, \mathbf{w}), i=1, \ldots, n]$\;
		$\mathbf{q} \leftarrow [\operatorname{auc}(i, \mathbf{w}), i=1, \ldots, n]$\;
	}
	\caption{Multi-armed bandit selection.}
	\label{algo:mab}
\end{algorithm}
First, the number of times each $k$ arm is selected $\mathbf{n} = [n_1~\cdots~n_k]$, the credit each arm obtained so far $\mathbf{q}=[q_1~\cdots~q_k]$, and the window of rewards $\mathbf{w}$ are all initialized. Then, we iterate over each selection step until the decision process ends. At first, if one or more arms have never been selected, we select randomly among these. Otherwise, we select an arm using the UCB formula, where $C$ is a parameter to control exploration. Next, we apply the selected arm and retrieve the reward. A tuple made of the arm index and the reward is prepended to the window of rewards, from which the oldest tuple is removed if we have reached the maximum window size $W$. Finally, the number of times each arm is selected is updated according to the window and the credit is given by the AUC (Algo.~\ref{algo:auc}) of each arm.

\section{Selecting Species}
\label{sec:popsel}

The main idea at the source of the proposed algorithm is very simple. Instead of evolving every population at each generation, as presented in Sec.~\ref{sec:coopcoev}, we want to choose the most promising population to evolve with the dynamic MAB described in Sec.~\ref{sec:mab}. This strategy has two major goals: it allows sustainable coevolution of the multiple subpopulations while balancing the global evolution resources allocated to each population, as they are not appearing in the evolution at the same time given the species addition and removal mechanisms used.

The evolution process starts similarly to the original algorithm. A number of species are initialized and one individual in each of them is selected as representative. The MAB is initialized with one arm associated with each species in the coevolution. Then, line~\ref{line:for} of Algo.~\ref{algo:coev} becomes lines~\ref{line:select} to \ref{line:endselect} of Algo.~\ref{algo:mab}, where the MAB selects the index $i$ of the species to be evolved. The application of variation operators, evaluation, and selection of $\mathfrak{S}_i$ remain the same. Later, the best individual of $\mathfrak{S}_i$ replaces the previous representative of species $i$ in $\mathfrak{R}$. The bandit is rewarded for the progress made by the representatives on the problem, and the rest of the iteration of Algo.~\ref{algo:mab} is run to complete the bandit update. The reward scheme will be detailed in the following Sec.~\ref{sec:reward}. Finally, the cooperative process continues with the improvement verification of the standard algorithm (lines~\ref{line:improvement} to \ref{line:endimprovement} of Algo.~\ref{algo:coev}). The improvement length is set to the value in the original algorithm times the number of species, since we evolve only one species on each generation. The arms in the bandit are adjusted for each addition and removal of a species in the coevolution, as detailed later in Sec.~\ref{sec:addarm}.

\subsection{Reward}
\label{sec:reward}

Properly rewarding the selected arm is a crucial element that should be carefully designed to ensure success. The choice of the reward must encourage exploitation of the best arm while still allowing exploration of the other arms. The concept used in our case is borrowed from an assumption made in the original UCB, contrary to what is presented by Fialho et al.~\cite{Fialho2010}. The reward given to the bandit is binary. An arm is assigned a reward of 1 if the fitness of the selected species representative is better than the one at the previous iteration, and 0 otherwise.

We argue that this reward method is the best that we can use in the context of population selection in cooperative coevolution compared to the direct and delta fitness assignment of Fialho et al.~\cite{Fialho2010}. On the one hand, the direct fitness allocation suffers when the bandit is updated with a new fitness that caused a loss -- if this fitness is still better than any other in the window that caused a gain, the arm will receive a higher credit for that loss than the other arm for its gain. This drawback has been addressed in the delta fitness assignment. On the other hand, this previous reward attribution is handicapped by the ``log convergence'' of optimization algorithms. The difference of fitness between consecutive iterations should decrease logarithmically with time. Sorting the delta fitness results in older data being privileged over new ones, which is counter intuitive as the current state of cooperation is probably closer to the next state than any older one. The Boolean reward allotment benefits from the properties of both assignments without their drawbacks, which explains its choice for the proposed method.

\subsection{Adding and Removing Arms}
\label{sec:addarm}

As species are added and removed from the evolution, the number of arms must change. Instead of entirely restarting the bandit, we simply modify its internal state. When adding a species, we add entries in both the number of times each arm has been selected and the credit each arm received so far, that is $\mathbf{n} \leftarrow [n_1~\cdots~n_k~0]$ and $\mathbf{q} \leftarrow [q_1~\cdots~q_k~0]$. This procedure forces the bandit to select the newly introduced species a couple of times, giving it a chance to find a suitable niche without being bumped out of the coevolution.

Removing an arm is conducted in a similar manner to the addition, with entries corresponding to the removed species in $\mathbf{n}$ and $\mathbf{q}$ being removed, leaving them with $k-1$ values each. For the window of reward $\mathbf{w}$, we remove all entries in which the removed species (indexed $l$) is involved, $\mathbf{w}~\leftarrow~[\forall (reward, arm) \in \mathbf{w} | arm\neq l]$.

\section{Experiments}
\label{sec:Experiments}

The proposed algorithm (CCEA-MAB) is tested and compared against the original cooperative coevolution (CCEA) of Potter and De Jong~\cite{Potter2001} on a string covering benchmark and a sensor placement problem.

\subsection{String Covering}
\label{sec:StringCovering}

The binary string covering problem,
introduced in \cite{Forrest1993} and used by \cite{Potter2001,Smith1993,Wu2010}, benchmarks cooperative evolutionary algorithms. The problem consists in matching a number of strings in a test set $\mathcal{T}$ with strings in a match set $\mathcal{M}$, where $\mathcal{T}$ is generally much larger than $\mathcal{M}$. Thus, the strings in the match set must generalize the patterns in the target set to obtain an optimal cover. The match strength $s(\mathbf{x},\mathbf{y})$ between two strings $\mathbf{x}$ and $\mathbf{y}$ is defined by the number of bits at the same position and of the same value that the two strings share. The strength of a match set $S(\mathcal{M},\mathcal{T})$ is the average maximum match strength its strings has against the target set,
\begin{equation*}
	S(\mathcal{M}, \mathcal{T}) = \frac{1}{|\mathcal{T}|} \sum_{\mathbf{t}_i\in\mathcal{T}} \max_{\mathbf{m}_j\in\mathcal{M}} s(\mathbf{m}_j, \mathbf{t}_i).
	\label{eq:mstrength}
\end{equation*}
The cooperative algorithm evolves a match set that is evaluated against the fixed test set. A number of arrangements of the fixed and variable bits is predetermined, these arrangements are called schemata. The test set consists of an equal number of each schemata where the variable bits have been replaced by random bit values. In all experiments each set strings are 64 bits long.

The representation used for this problem is a 64 bits string. We use a two-point crossover and a flip bit mutation to generate the offspring, each variation operator being applied with a certain probability. The fitness of an individual is given by the match strength of the group it forms with the representatives of the other species. Tournament selection is used to select individuals in each species. The contribution of a species is computed by the number of times its representative has the best cover for a target string.

Four versions of the string covering problem are used to assess each property of the cooperative evolutionary algorithm of Potter and De Jong~\cite{Potter2001}. While the two first tests gauge the ability of the algorithm to locate and cover multiple niches, and to evolve the appropriate level of generality, we are more interested in the capacity of adaptation to a changing number of species and the ability to evolve the right number of species. The parameters of the algorithms are shown in Tab.~\ref{tab:param_string}. The parameters concerning the evolutionary algorithms and the coevolutionary process are set similar to what is presented in Potter and De Jong~\cite[pp. 10 and 17]{Potter2001} and those for the bandit are set accordingly to the conclusions made by Fialho~\cite{Fialho2010}.
\begin{table}
	\centering
	\begin{tabular}{l|c|c} \hline
		Parameter						& CCEA	& CCEA-MAB\\ \hline
		Species Size					& 50		& 50\\
		Initial Number of Species				& 1		& 1\\
		Crossover Rate				& $0.6$	& $0.6$\\
		Mutation Rate					& $1.0$	& $1.0$\\
		Flip Bit Rate						& $1/64$	& $1/64$\\
		Tournament Size				& 3		& 3\\ \hline
		Window Size ($W$)			& --		& 50	\\
		Decay Factor ($d$)			& --		& $1.0$\\
		Exploration Factor ($C$)	& --		& $1.0$\\ \hline
		Improvement Length ($I$)& 5		& 5\\
		Improvement Threshold ($T_\mathrm{i}$)	& $0.5$	& $0.5$\\
		Extinction Threshold ($T_\mathrm{c}$)	& $5.0$	& $5.0$\\ \hline
	\end{tabular}
	\caption{Parameters used in the experiments.}
	\label{tab:param_string}
\end{table}

\subsubsection{Adaptation During the Evolution}

The capability of adaptation of our version of the coevolutionary algorithm is crucial. As more time is allocated to more promising species, other species must still have enough time to adapt their representatives to the addition of new species. This test allows the ability of specialization of the species over time to be measured as more and more species are added to the evolution.

For this experiment, the evolution is not in charge of adding or removing species, allowing to focus solely on the adaptation property of the algorithm. Starting with a single species, a new species is added at each 100 generations until there is the same number of species as schemata to be matched. At the end of the evolution all schemata should be covered. Three scenarios have been designed. 

\noindent\textbf{Scenario 1} is similar to the one used by Potter and De Jong, where a 30 element target set is generated from the three following schemata, each containing 32 fixed bits and 32 variable bits.\\

\noindent\texttt{\centering
1\#\#1\#\#\#1\#\#\#11111\#\#1\#\#1111\#1\#\#1\#\#\#1\#1111\#\#111111\#\#1\#11\#1\#11\#\#\#\#\#\#\\
1\#\#1\#\#\#1\#\#\#11111\#\#1\#\#1000\#0\#\#0\#\#\#0\#0000\#\#000000\#\#0\#00\#0\#00\#\#\#\#\#\#\\
0\#\#0\#\#\#0\#\#\#00000\#\#0\#\#0000\#0\#\#0\#\#\#0\#0000\#\#001111\#\#1\#11\#1\#11\#\#\#\#\#\#\vspace{2ex}\\
}

\noindent\textbf{Scenario 2} is similar to the first one, with a 30 element target set. Five schemata are generated randomly to produce the target set, using the same noise pattern (variable bits) as in scenario 1, while the fixed bits correspond to those of a random integer uniformly chosen in $[0, 2^{32}-1]$. 

\noindent\textbf{Scenario 3} is composed of a 50 element target set generated from five schemata, with a varying number of variable bits. For each schema, we first draw uniformly from $[16, 48]$ the number of variable bits ($v$) used. Then, the fixed bits are generated using the binary representation of a number taken uniformly in $[0, 2^{64-v}-1]$. Finally, the variable and fixed bits are shuffled together to form a 64-bit schema.

It should be stressed that for all scenarios, all runs of CCEA and CCEA-MAB are carried out in pairs, with the same target set used for each pair of runs. This allows a fair comparison between the methods.

\begin{figure}[t]
	\centering
	\subfloat[CCEA]{
		\includegraphics[width=0.70\linewidth]{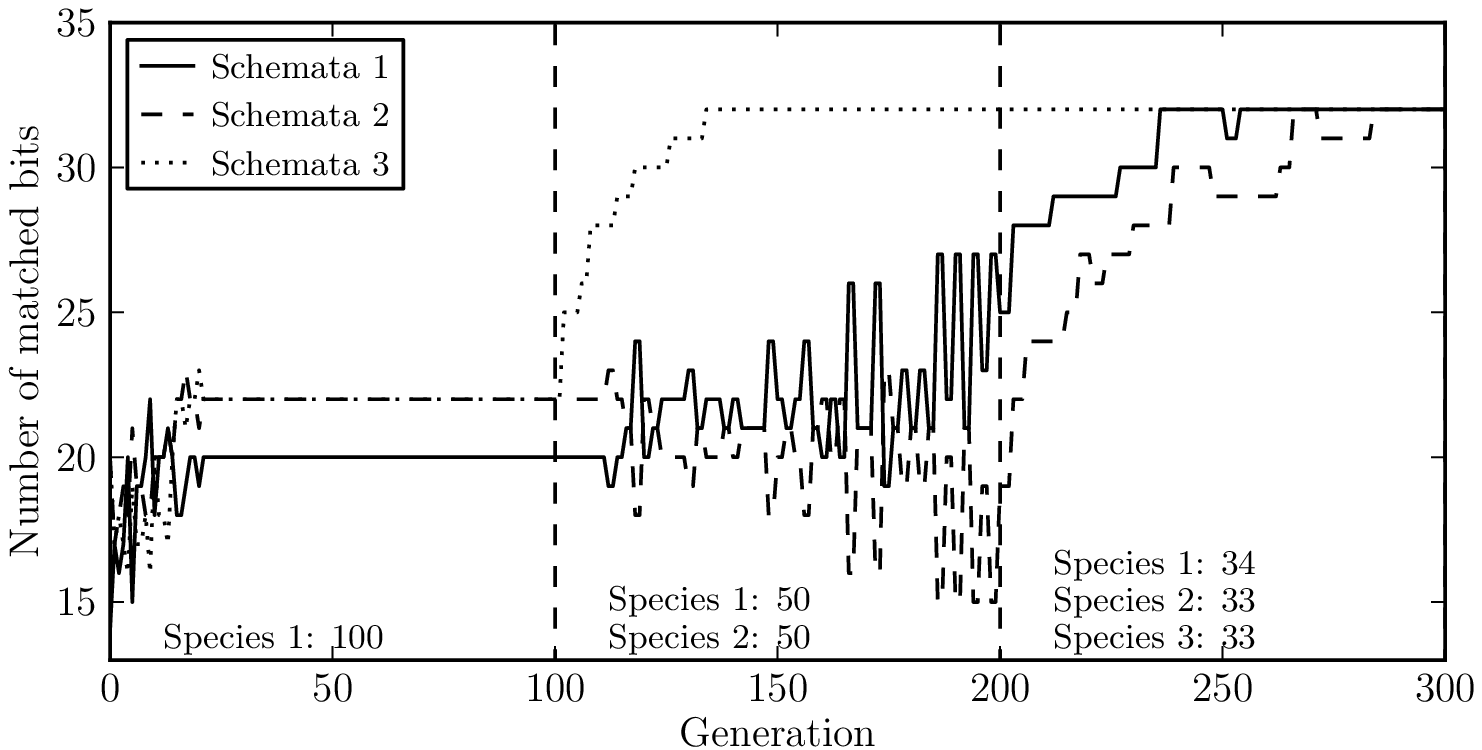}
	}
	
	\vspace{-1em}
	\subfloat[CCEA-MAB]{
		\includegraphics[width=0.70\linewidth]{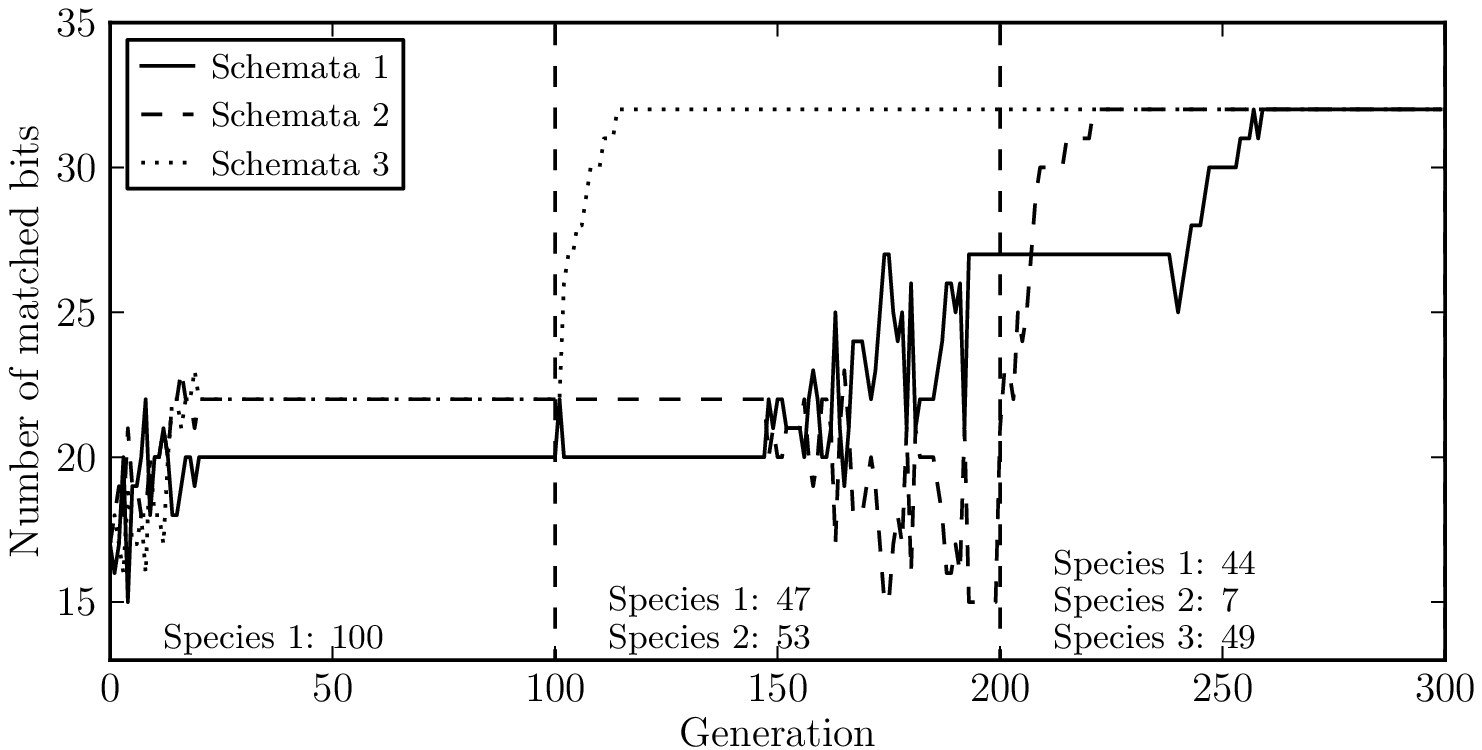}
		\label{fig:adapt_mab_orig}
	}
	\caption{Maximum number of matched bits by a single match string on each schemata in scenario 1 with the number of times each species is evolved.}
	\label{fig:adapt_basic}
\end{figure}

On 100 runs for scenario 1, the original and the MAB-driven algorithms find the optimal string cover on all occasions. A typical run is shown in Fig.~\ref{fig:adapt_basic}, where the match strength against each schemata is presented.
Fig.~\ref{fig:adapt_basic}\subref{fig:adapt_mab_orig} clearly illustrates the allocation to the promising species right after generation 100 and 200 where the slope of the number of bits matched is generally steeper for CCEA-MAB. We also notice that once a species has converged to a schemata, time is allocated to adjust the other species, if there can be a reward, as exposed by the difference between the number of times each species is evolved in CCEA-MAB. In the third part of Fig.~\ref{fig:adapt_basic}\subref{fig:adapt_mab_orig}, we observe that species 2, which has converged to schemata 3, is chosen by the bandit only 7 times over the 100 last generations while the two other species for which convergence is still possible are chosen 44 and 49 times. This process saves much effort since the progress made by evolving a species is not interrupted by evolving another less promising or already converged one.

For the second scenario with five schemata, CCEA finds the optimal cover in only $22\%$ of the experiments, while CCEA-MAB succeeds $71\%$ of the time.
\begin{figure}[t]
	\centering
	\subfloat[CCEA]{
		\includegraphics[width=0.70\linewidth]{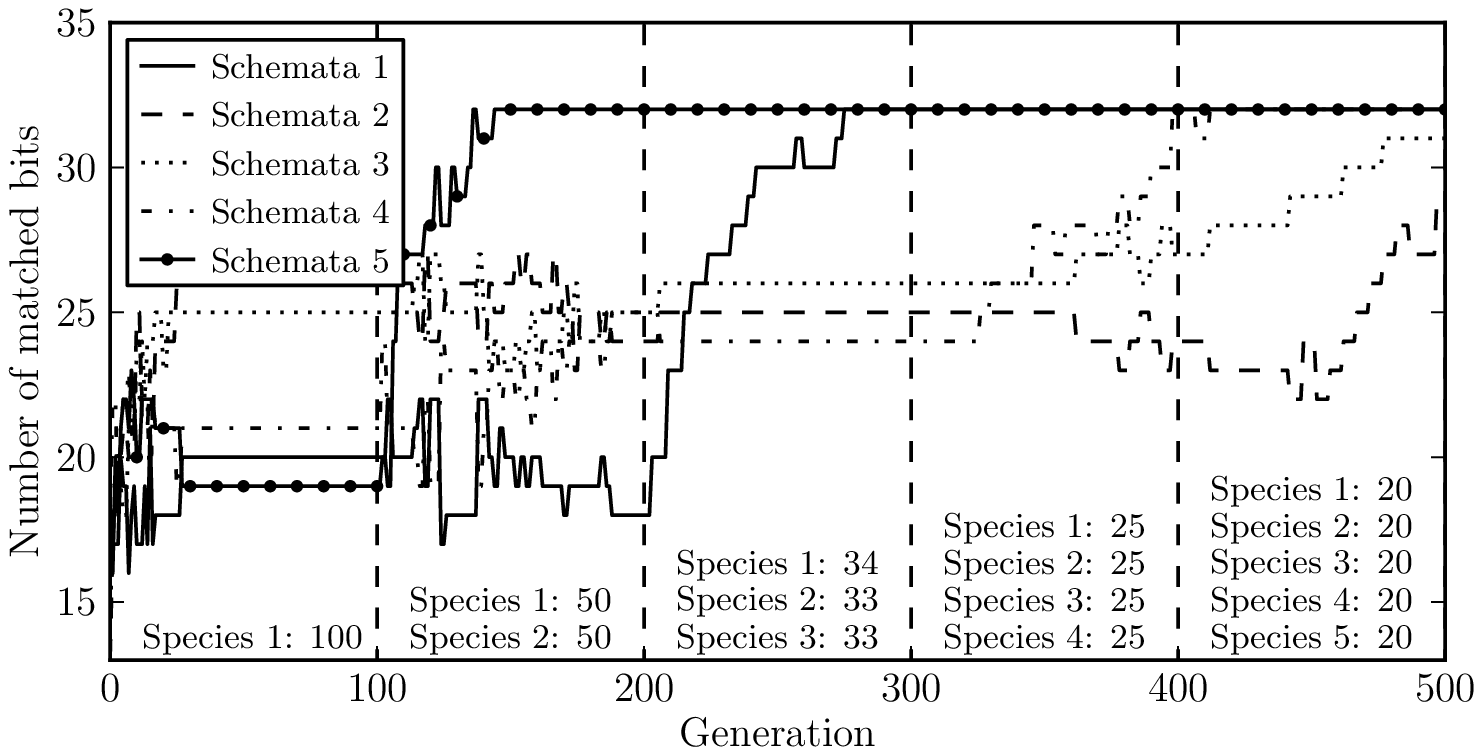}
		\label{fig:adapt_orig_5}
	}
	
	\vspace{-1em}
	\subfloat[CCEA-MAB]{
		\includegraphics[width=0.70\linewidth]{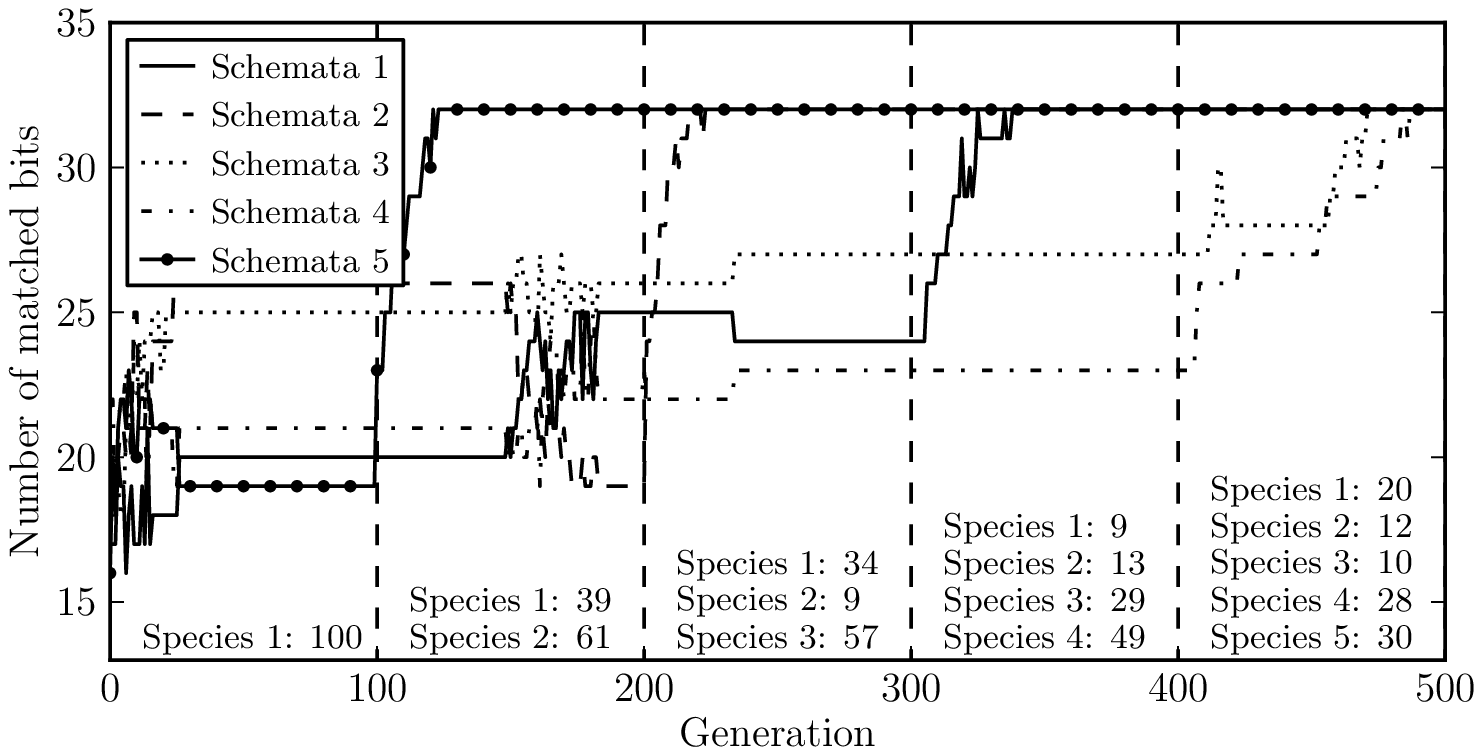}
		\label{fig:adapt_mab_5}
	}
	\caption{Maximum number of matched bits by a single match string on each schemata in scenario 21 with the number of times each species is evolved.}
	\label{fig:adapt_5}
\end{figure}
Fig.~\ref{fig:adapt_5} exposes a typical run with this configuration. We see that the two algorithms have mostly the same behaviour for the first couple of phases, but when the fourth and fifth species are introduced, alternating between the species in a round-robin fashion does not leave enough generations to the important species to converge to a niche, as seen in Fig.~\ref{fig:adapt_5}\subref{fig:adapt_orig_5}. The greater number of generations offered to the promising populations by CCEA-MAB in the fourth and fifth phases greatly helps to find an optimal coverage (Fig.~\ref{fig:adapt_5}\subref{fig:adapt_mab_5}).

The experiments with scenario 3 show a success rate of $83\%$ for CCEA and $88\%$ for CCEA-MAB. While this appears to be contradictory with the previous experiment, we find that the large amount of time allocated between the introduction of new species (100 generations) greatly helps CCEA while not affecting the MAB version much. In fact, just reducing the number of generations allocated per species introduction to 75 reduces the number of perfect coverages found by CCEA to $10\%$ and those found by CCEA-MAB to $33\%$, conserving its superiority observed previously.

\subsubsection{Finding the Appropriate Number of Species}

In this problem, the coevolutionary algorithm aims not only at evolving the perfect cover for a given target set, but also strives at finding the optimal number of species for that problem. For this test, we expect the MAB version of the coevolutionary algorithm to perform much better, with less species removed of their inability to identify a suitable niche in the allocated time. This test is the closest one to a real world problem where we do not know how many species will be required, nor the time we should allocate after the addition of a species for the algorithm to stabilize.

For this experiment, the complete algorithms presented in Sec.~\ref{sec:coopcoev} and \ref{sec:popsel} are used. The coevolution must concurrently determine the appropriate number of species, a niche for all those species, and the optimal coverage for the problem. Again starting with a single species, the cooperative coevolution mechanisms will be used to add and remove species. The evolution is run until the algorithm finds the perfect coverage or 500 generations have passed. We will use the same three scenarios as in the adaptation experiment.

\begin{figure}
	\centering
	\includegraphics[width=0.70\linewidth]{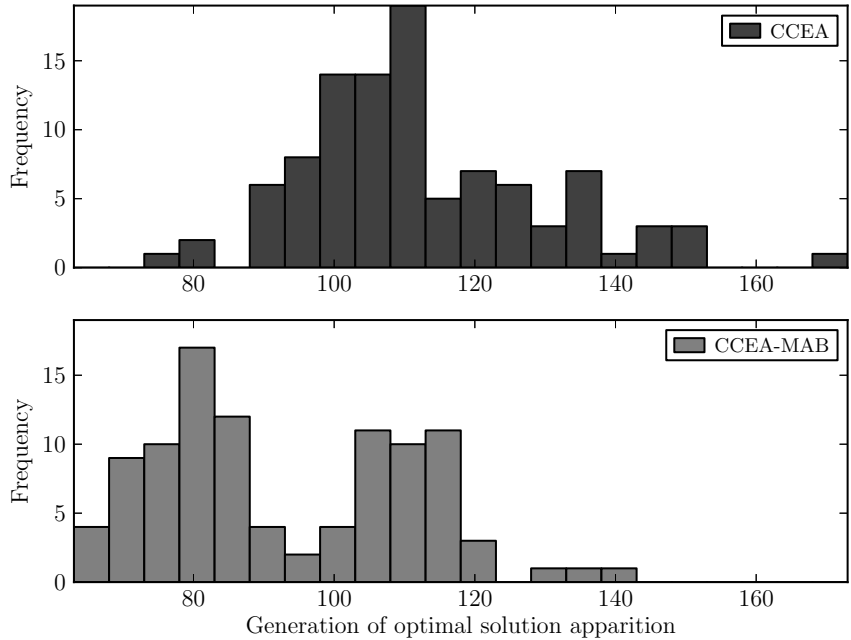}
	\caption{First hit generation histogram for scenario 1.}
	\label{fig:evol_basic_hist}
\end{figure}

Fig.~\ref{fig:evol_basic_hist} shows the generation at which the perfect coverage was found for scenario 1. The average first hit generation is $111.9$ for CCEA and $92.5$ for CCEA-MAB.
A Wilcoxon signed test reveals that the two algorithms are different with a significance level of $99.9\%$. 

\begin{figure}[t]
	\centering
	\includegraphics[width=0.70\linewidth]{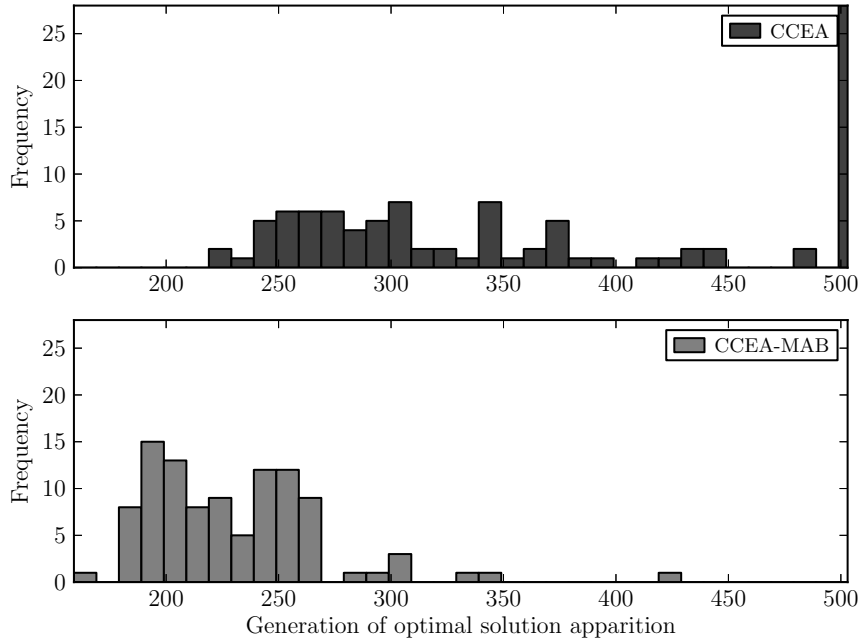}
	\caption{First hit generation histogram for scenario 2.}
	\label{fig:evol_5_hist}
\end{figure}

\begin{figure}[t]
	\centering
	\subfloat[CCEA]{
		\includegraphics[width=0.70\linewidth]{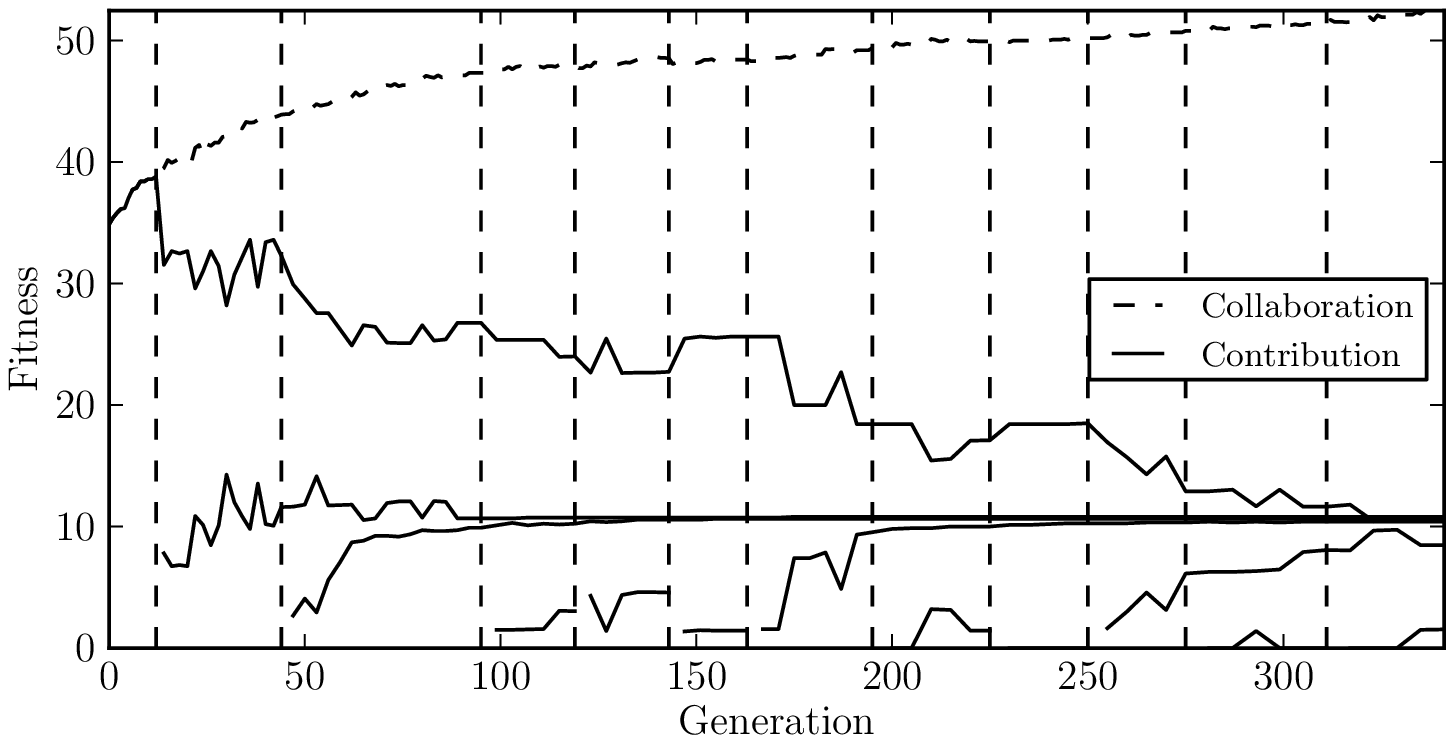}
		\label{fig:evol_orig_5}
	}
	
	\vspace{-1em}
	\subfloat[CCEA-MAB]{
		\includegraphics[width=0.70\linewidth]{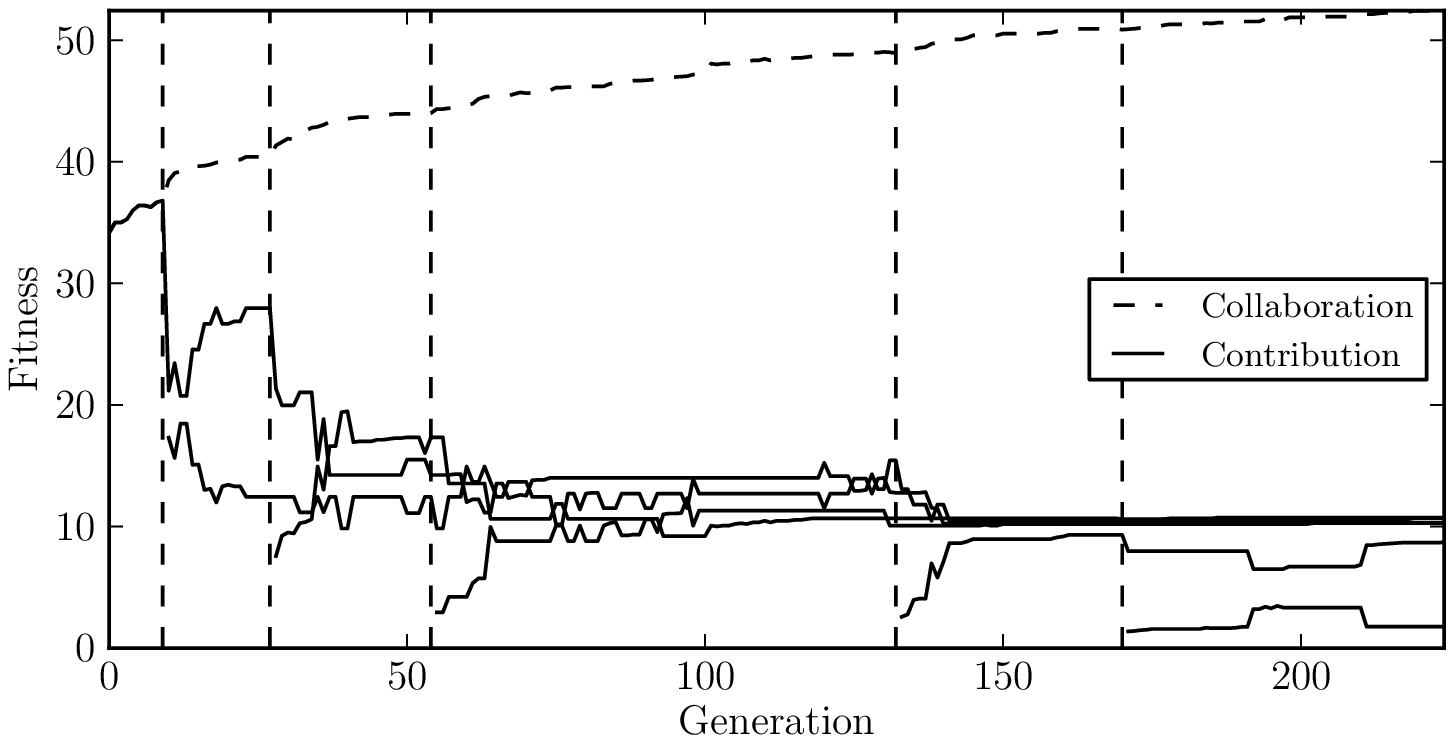}
		\label{fig:evol_mab_5}
	}
	\caption{Collaboration (match set fitness) and contribution (fitness of each species independently) for scenario 2.}
	\label{fig:evol_5}
\end{figure}

The first hit generation histogram for scenario 2 is shown in Fig.~\ref{fig:evol_5_hist}.
The average convergence time for CCEA, excluding those exceeding the 500 generations allocated, is $315.0$ generations, while it is $229.7$ generations for CCEA-MAB. A Wilcoxon signed test indicates that the two algorithms are different at a significance level of $99.9\%$.
We also notice that CCEA-MAB does find the optimal coverage on all of the 100 runs, while CCEA does not converge in more than $25\%$ of the experiments. Fig.~\ref{fig:evol_5} presents the run with the median first hit time for each algorithm. The contribution of the different species is shown with the continuous lines and the collaboration fitness is presented with the dashed line.
This clearly illustrates that CCEA consumes much more species than CCEA-MAB. In fact, it takes 12 species and approximately 350 generations for CCEA to have the right elements to entirely cover the target set, while only 6 species and less than 250 generations are required for CCEA-MAB. This clearly shows that allocating more steps to the promising species allows them to find suitable niches and support global improvements in the optimization, while the original algorithm would have removed those species at the outset.

\begin{figure}[t]
	\centering
	\includegraphics[width=0.70\linewidth]{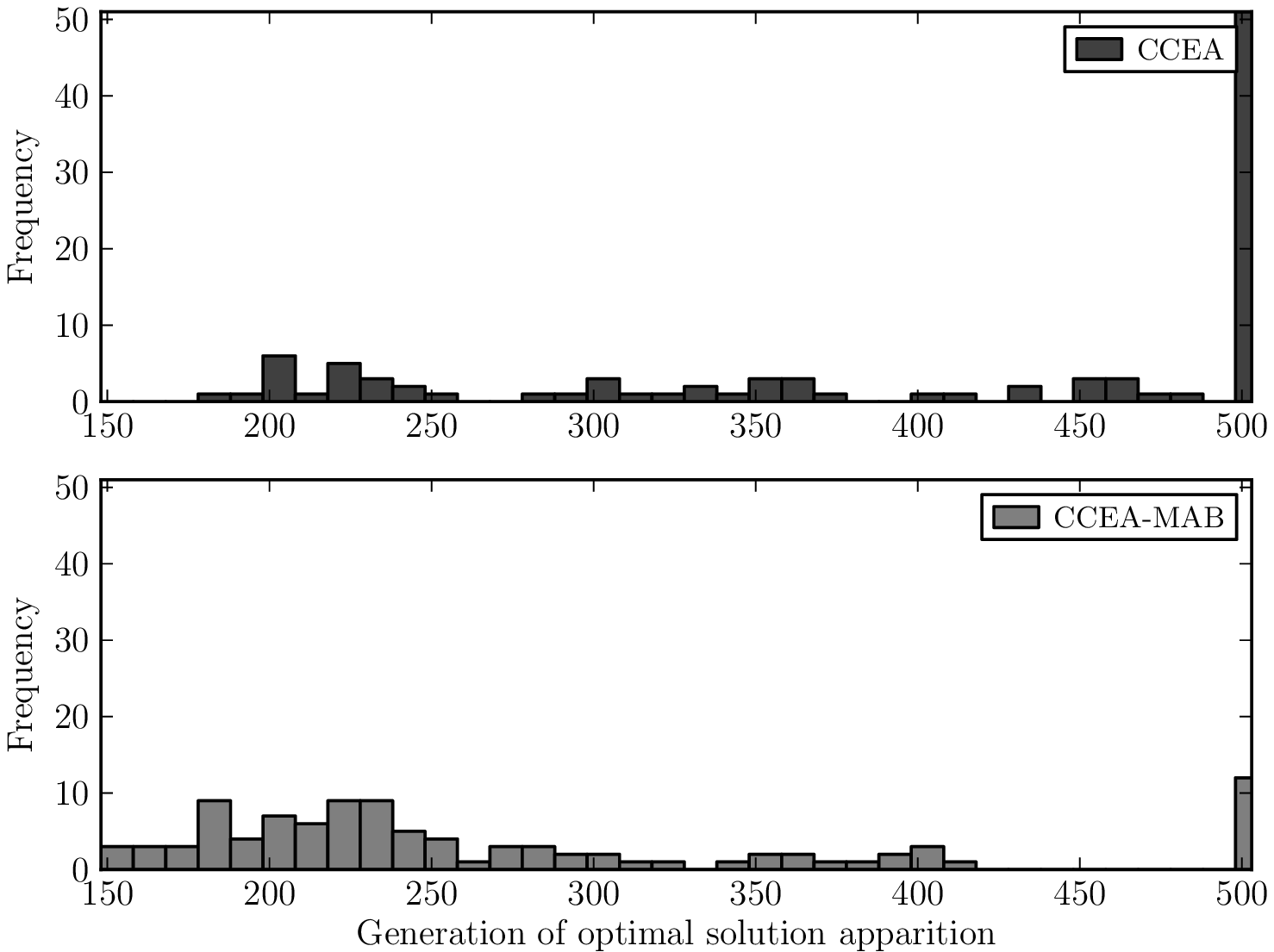}
	\caption{First hit generation histogram for scenario 3.}
	\label{fig:evol_rand5_hist}
\end{figure}

\begin{figure}[t]
	\centering
	\subfloat[CCEA]{
		\includegraphics[width=0.70\linewidth]{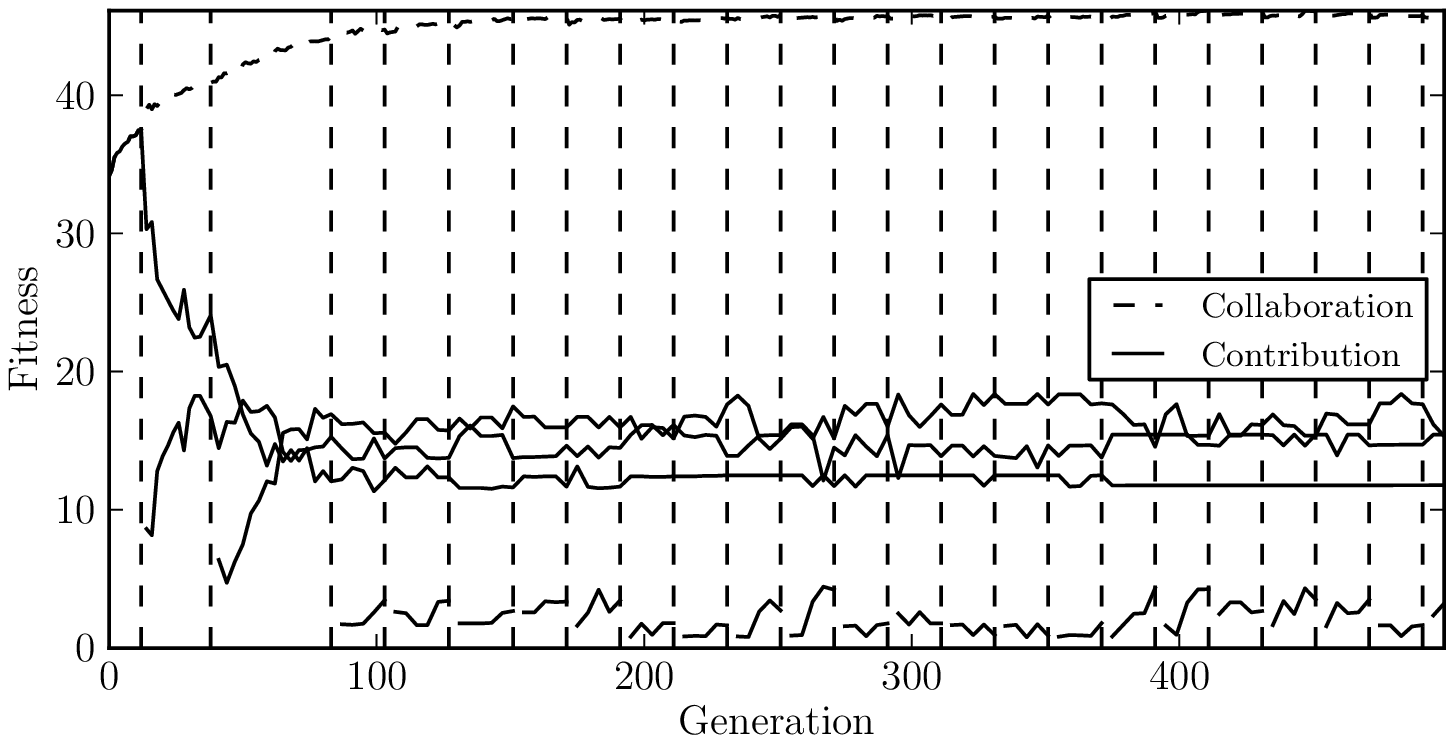}
		\label{fig:evol_orig_rand5}
	}
	
	\vspace{-1em}
	\subfloat[CCEA-MAB]{
		\includegraphics[width=0.70\linewidth]{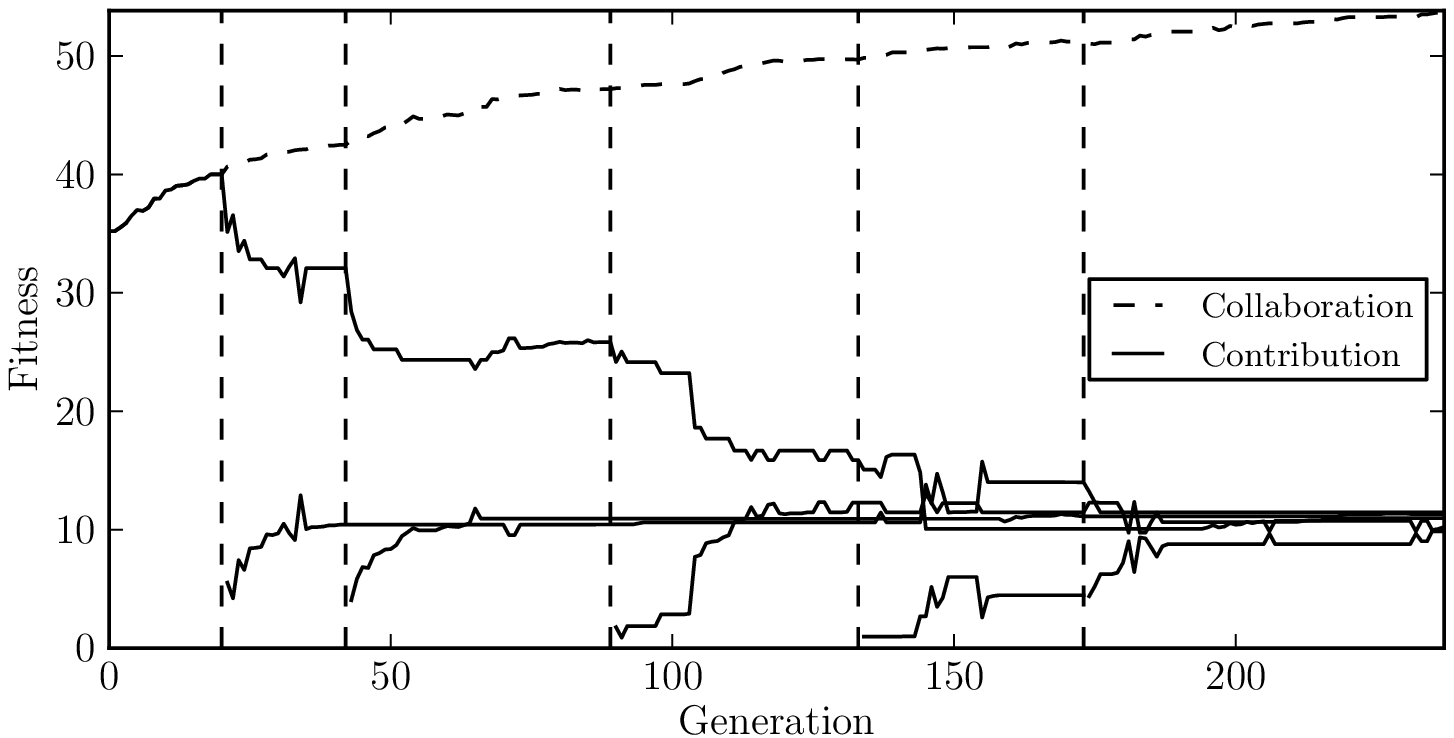}
		\label{fig:evol_mab_rand5}
	}
	\caption{Collaboration (match set fitness) and contribution (fitness of each species independently) for scenario 3.}
	\label{fig:evol_rand5}
\end{figure}

Finally, the first hit generation histogram for scenario 3 is shown in Fig.~\ref{fig:evol_rand5_hist}. 
The average convergence time for CCEA  and CCEA-MAB on this experiment, excluding those exceeding the 500 generations allocated, are respectively $313.8$ and $245.6$ generations. The runs that have not converged represent more than $50\%$ of the trials for CCEA and only $11\%$ of the runs for CCEA-MAB. Again, the Wilcoxon signed test indicates that the two algorithms are different at a significance level of $99.9\%$ in this configuration.
Fig.~\ref{fig:evol_rand5} exposes the same problem as in the previous scenario, showing the runs corresponding to the median first hit time of each algorithm. 
We see that in CCEA, an enormous amount of evaluations are wasted on species immediately removed after their introduction. However, these species should have helped the algorithm to increase its collaboration fitness. This problem is not present in CCEA-MAB as it allocated enough resources to the promising species.

\subsection{Sensor Placement}
\label{sec:SensorPlacement}

The string covering problem exposed the properties of our new algorithm. This section will illustrate how this algorithm performs on a more complex case.
As shown by De Rainville et al.~\cite{DeRainville2012}, cooperative coevolutionary algorithms apply well to the sensor placement optimization. We will study the effect of selecting the species to evolve on each generation on a similar scenario.

The sensor placement problem we face is to optimize the position of a group of sensors (i.e. directional cameras) so that an entire section of an environment is seen. The required number of sensors is unknown, thus the adaptation of the number of species in the evolution is essential. The representation used to tackle this problem is a real-valued vector representing the position and orientations of a sensor. Each species will evolve a single sensor and cooperation between species will produce multiple sensor solutions. We use a simulated binary crossover and a Gaussian mutation as variation operators. Selection between the individuals of a species is achieved by a tournament. The visibility of a sensor is computed by the integral of its resolution on the surface it senses. For space reasons, the details on computing a sensor visibility cannot be entirely exposed here.

\begin{figure}[t]
	\centering
	\includegraphics[width=0.35\linewidth]{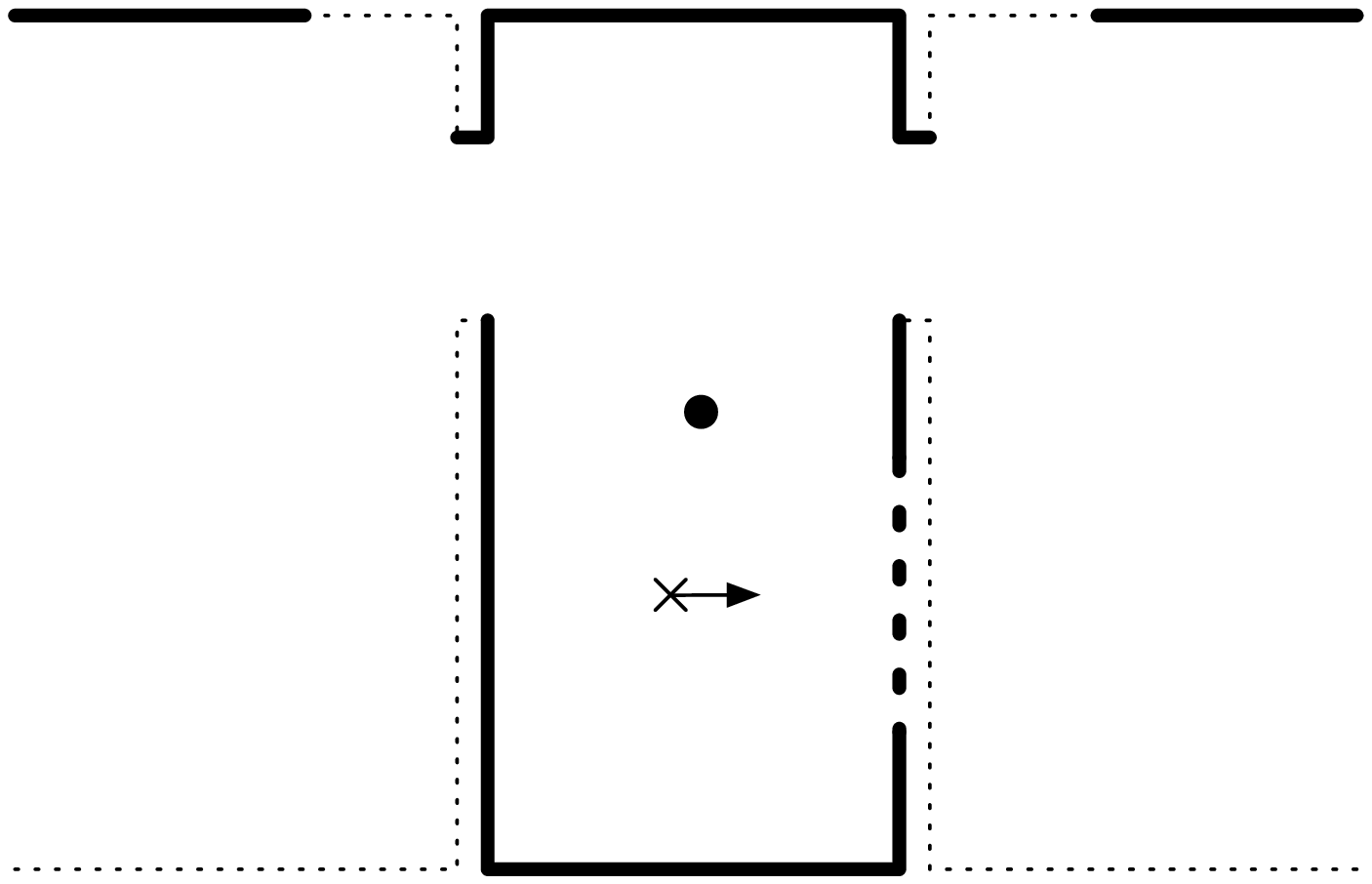}
	\caption{The environment used for the real world experiments. The desired view and the sensor view are shown respectively by the thick continuous and thick dashed lines.}
	\label{fig:visibility}
\end{figure}

The collaboration between species is evaluated by computing the complete visibility attained with all representatives. The fitness is the difference between the visibility required for the environment and the visibility given by the group of sensors. Fig.~\ref{fig:visibility} shows the different visibility concepts implied in the sensor placement problem. In this figure, the environment desired view is determined by the resolution an omnidirectional camera would have of the environment if it was positioned on the dot.
The visibility of a sensor is computed using the same process, a sensor placed at position ``x'' and oriented as shown by the arrow sees the thick dashed line. The fitness is the integral of the difference between the required visibility and the visibility of the sensor. Finally, a species contribution is computed as the difference of fitness the group of representatives would have with and without the individual from that species.

\begin{figure}[t]
	\centering
	\includegraphics[width=0.70\linewidth]{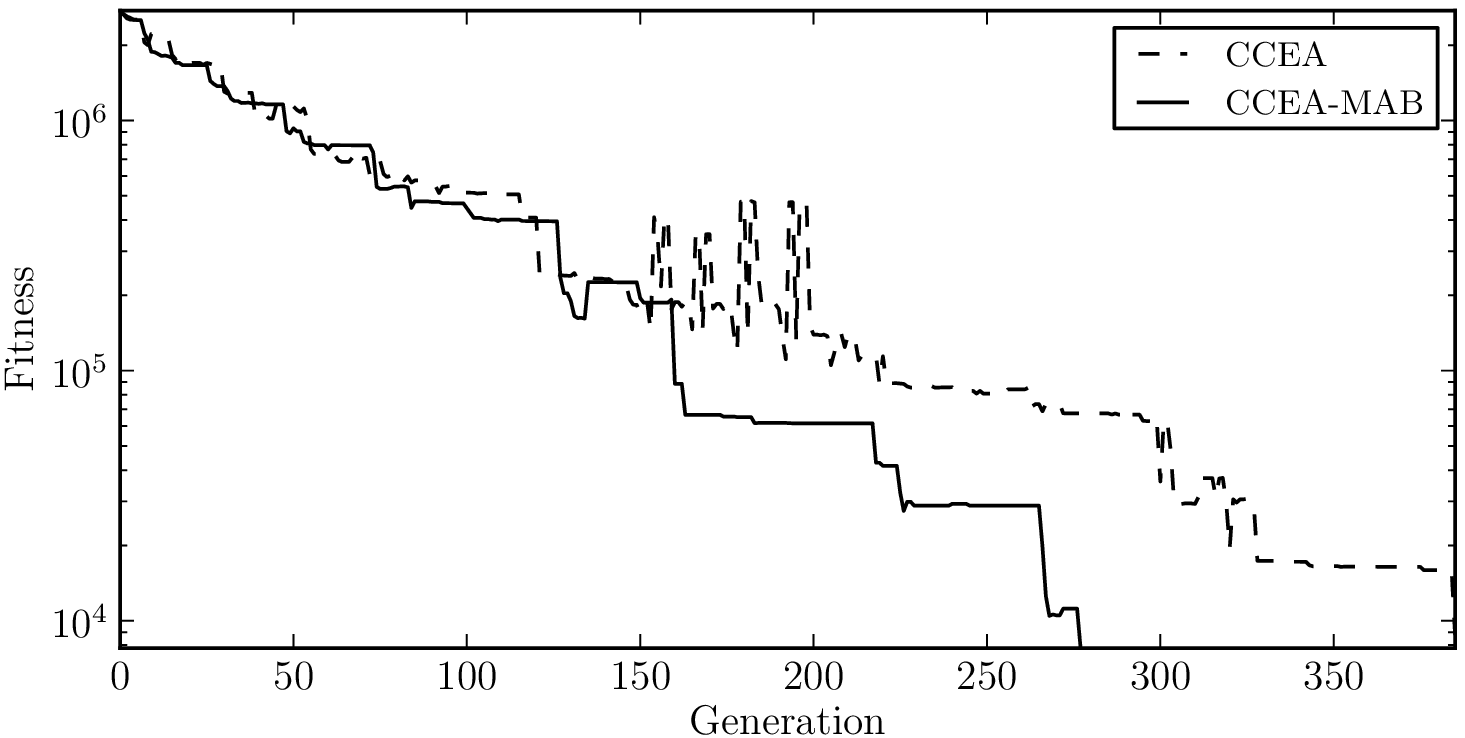}
	\caption{Fitness progress on the sensor problem.}
	\label{fig:sensor_evol}
\end{figure}

Experiments have been conducted for the environment of Fig.~\ref{fig:visibility}.
The desired field of view is $360^\circ$ while the sensors field of view is $45^\circ$, which results in a theoretical optimal solution with 8 sensors. We stopped the evolution for both techniques when a suitable coverage had been achieved, namely an error less than $10^4$. The average first hit generation and average resulting number of sensors, for 10 independent runs, are shown in Tab.~\ref{tab:results_sensors}.
\begin{table}[t]
	\centering
	\begin{tabular}{l|c|c} \hline
		Algorithm	&  First Hit Gen.	& \# Sensors	\\ \hline
		CCEA		& $429.4$	& $8.8$	\\ 
		CCEA-MAB	& $283.7$	& $8.8$	\\ \hline
	\end{tabular}
	\caption{Results for the second environment.}
	\label{tab:results_sensors}
\end{table}
A Mann-Whitney-Wilcoxon test reveals that the two algorithms are different with a significance level of $99.9\%$. This clearly shows that the MAB maintained the focus of the optimization on the important sensors, and thus the algorithm converged more rapidly to a suitable solution. Fig.~\ref{fig:sensor_evol} shows the evolution of the collaboration fitness during the median run for both algorithms.
As mentioned in the previous section, when many species are present, the MAB greatly increases the convergence of the algorithm because it allocates resources to the preferred population and offers some stability to the new species. Moreover, tuning the extinction threshold would force the algorithm to reach a better position for each sensor and not use the extra sensors, but it would also require more computational time to identify that solution. In addition, a fixed length real valued genetic algorithm has been run for several experiments with different number of sensors without success at producing any suitable solution even with carefully tuned parameters.

\section{Conclusion}
\label{sec:Conclusion}

This paper presented an efficient way to automatically control the resources allocated to the cooperative populations in coevolution in order to sustain co-adaptation. Our technique exploits the complete potential of every species and learns when to select another species due to our multi-armed bandit. Our technique also alleviates the burden of precisely choosing the improvement length, as the allocation of the resources to the population is conducted automatically. Finally, we introduced a mechanism that responds to the difficulty of each problem by allocating the computing resources to the subpopulations which will sustain evolution. The string covering benchmark and the sensor placement problem clearly illustrate that our method helps the algorithm to converge more rapidly to a suitable solution in situations where the original algorithm struggles to keep up with the problem.

This work can be extended in a number of ways. First, different multi-armed bandits exist and much work has been carried out to solve combinatorial problems with these learners. It would be interesting to incorporate the dynamic technique in these bandits and use them in our algorithm as they do not make the independence assumption between the populations.
Second, preliminary results show that the population selection technique can be applied efficiently to other types of coevolution such as competitive coevolution. In this case, the bandit would balance the progress made by both competitors so that one does not entirely dominate the other. Third, the bandit technique would also be promising in multi-population algorithms to invest resources in the most promising areas of the search space.

\section*{Acknowledgements}
This work has been funded by the FRQNT (Qu\'ebec) and NSERC (Canada). The authors wish to thank Annette Schwerdtfeger for proofreading the manuscript.

\bibliographystyle{plain}
\bibliography{biblio}

\end{document}